\documentclass[10pt,twocolumn,letterpaper]{article}

\newcommand{\comment}[1]{}

\newcommand{\etal}{\emph{et al. }}

\newcommand{\bu}{\mathbf{u}}

\newcommand{\bx}{\mathbf{x}}

\newcommand{\bB}{\mathbf{B}}

\newcommand{\bI}{\mathbf{I}}

\newcommand{\bU}{\mathbf{U}}
\newcommand{\bV}{\mathbf{V}}

\newcommand{\bX}{\mathbf{X}}

\usepackage{colortbl}
\usepackage{array}

\usepackage{cvpr}
\usepackage{times}
\usepackage{epsfig}
\usepackage{graphicx}
\usepackage{amsmath}
\usepackage{amssymb}
\usepackage{rotating}
\usepackage{upgreek}
\usepackage{balance}
\usepackage{bbding}
\usepackage{hhline}
\usepackage{listings}

\usepackage{algorithm}
\usepackage{algorithmic}
\usepackage[utf8]{inputenc}

\usepackage[breaklinks=true,bookmarks=false]{hyperref}

\cvprfinalcopy 

\def\cvprPaperID{2967} 

\ifcvprfinal\fi
\begin{document}

\title{{Geometry-Aware Network for Non-Rigid Shape Prediction from a Single View}}

\author{A. Pumarola$^{1}$ \hspace{0.7cm}
A.  Agudo$^{1}$ \hspace{0.7cm}
L.  Porzi$^{2}$ \hspace{0.7cm}
A.  Sanfeliu$^{1}$ \hspace{0.7cm}
V.  Lepetit$^{3}$ \hspace{0.7cm}
F.  Moreno-Noguer$^{1}$ \\
$^{1}$Institut de Rob\`otica i Inform\`atica Industrial, CSIC-UPC, Barcelona, Spain\\
$^{2}$Mapillary Research, Graz, Austria\\
$^{3}$Laboratoire Bordelais de
Recherche en Informatique, Universit\'e de Bordeaux, France
}

\maketitle

\begin{abstract}

We propose a method for predicting the 3D shape of a deformable surface from a single view. By contrast with previous approaches, we do not need a pre-registered template of the surface, and our method is robust to the lack of texture and partial occlusions. At the core of our approach is a {\it geometry-aware} deep architecture that tackles the problem as usually done in analytic solutions: first perform 2D detection of the mesh and then estimate a 3D shape that is geometrically consistent with the image. We train this architecture in an end-to-end manner using a large dataset of synthetic renderings of shapes under different levels of deformation, material properties, textures and lighting conditions. We evaluate our approach on a test split of this dataset and available real benchmarks, consistently improving state-of-the-art solutions with a significantly lower computational time. 

\end{abstract}

\section{Introduction}

Motivated by the current success of deep learning methods for estimating a depth map from a single image of a scene~\cite{eigen2014depth, garg2016unsupervised,GodardCVPR2017}, in this paper we tackle the related problem of estimating the underlying parametric model defining the shape of a non-rigid surface from a single image. This problem has been traditionally addressed in the context of the Shape-from-Template (SfT) paradigm~\cite{bartoli2015shape}, requiring a reference template image of the surface for which the 3D geometry is known, and a set of 3D-to-2D point correspondences or a mapping between this template and the input image. This approach, however, may be difficult to hold in practice, specially when considering low-textured surfaces. 

In this work we relax previous assumptions and present a learning-based approach that allows for globally non-rigid surface reconstruction from a single image without relying on point correspondences, and which in particular, shows robustness to situations rarely addressed previously: lack of surface texture and large occlusions. Our model is based on a fully differentiable Deep Neural Network that estimates a 3D shape from a single image in an end-to-end manner, and builds upon three branches that enforce geometry consistency of the solution.

\begin{figure*}[t!]
	\centering
	\includegraphics[width=\textwidth]{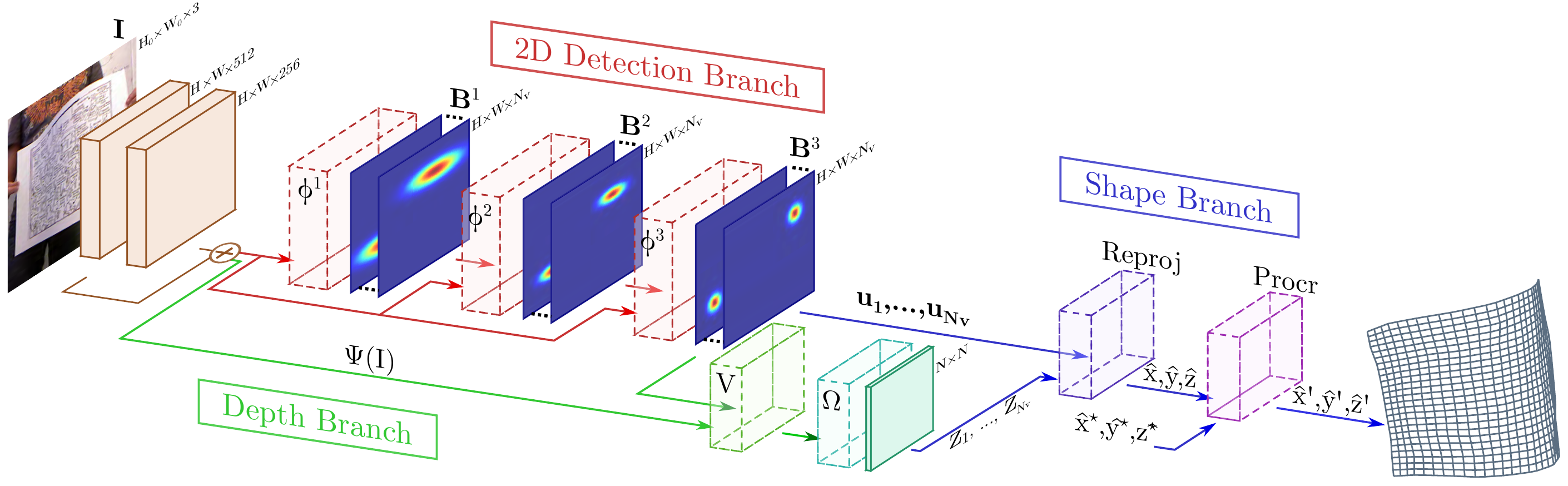}
	\caption{{\bf Overview of our approach.} The proposed architecture consists of three main branches. The `2D Detection Branch' is responsible for the 2D location of the mesh and the associated belief maps. The `Depth Branch' lifts the 2D detected mesh by leveraging on image cues and the detection uncertainties. Finally, the `Shape Branch' fuses the 2D detections and their estimated depths to obtain 3D shape in such a way  that perspective projection is enforced. An additional `Procrustes Layer' is used during training to align the estimated mesh with the ground truth one.}
	\label{fig:network}
\end{figure*}

More exactly, as illustrated in Fig.~\ref{fig:network}, a first branch of the proposed architecture (the `2D Detection Branch') is responsible for localizing the mesh onto the image, and for fitting a 2D grid to it. The 2D vertices of this grid are then lifted to 3D by the `Depth Branch', a regressor that combines the 2D detector confidence maps and the input image features. Finally, a `Shape Branch' is responsible for recovering the full shape while ensuring that the estimated 3D coordinates correctly re-project onto the image. During training, this branch also incorporates a novel fully-differentiable layer that performs a Procrustes transformation and aligns the estimated 3D mesh with the ground truth one. This branch is important as it was proven important to perform Procrustes alignment in previous approaches for adapting to datasets with different reference frames and metrics. It also favors convergence of the learning process. 

Since there is no dataset large enough to train data-hungry deep learning algorithms such as ours, we have created our own using a rendering tool. We have synthesized 128,000 photo-realistic pairs input 2D-image/3D-shape accounting for different levels of deformations, amount and type of texture, material properties, viewpoints, lighting conditions and occlusion. Figure~\ref{fig:syntheticresults}-Top shows some examples. Evaluation on a test split of this dataset demonstrates remarkable improvement of our network compared to state-of-the-art SfT techniques, which typically rely on known 3D-to-2D correspondences, especially under strong occlusions and poorly-textured surfaces. Furthermore, our model learned with synthetic data can be easily fine-tuned to real sequences, using just a few additional real training samples. Results on the CVLab sequences~\cite{varol2012constrained} with a bending paper and a deforming t-shirt again clearly show that our method outperforms existing approaches.

In summary, our main contributions are: 1) the first---to the best of our knowledge---fully-differentiable model for non-rigid surface reconstruction from a single image that does not require initialization,   accurate knowledge of the template, 3D-to-2D correspondences, nor hand-crafted constraints; 2) a geometry-aware architecture that embeds a pinhole camera model and encodes rigid alignment during training; and 3) a large photo-realistic dataset of images of non-rigid surfaces annotated  with the corresponding 3D shapes, which will be made publicly available, and we hope it will inspire future research in the field.

\section{Related Work}
Reconstructing non-rigid surfaces from monocular images is known to be a severely ill-posed problem which requires introducing different sources of prior knowledge in order to be solved. In this section, we will split related work into methods that define these priors based on pre-defined models (either physically-based or handcrafted) and techniques that learn them from training data.

Early approaches described non-rigid surfaces using models inspired by physics, such as superquadrics~\cite{metaxas1991constrained}, thin-plates~\cite{mcinerney1993finite}, elastic models~\cite{kita1996elastic} and finite-elements~\cite{mcinerney1995dynamic}. These representations, however, could not accurately approximate the non-linear behavior of large deformations. 

More complex deformations can be captured by SfT approaches~\cite{bartoli2015shape,chhatkuli2014stable,moreno2013stochastic,moreno2009capturing,ostlund2012laplacian,perriollat2011monocular, salzmann2009reconstructing,salzmann2008closed,vicente2012soft}, which aim at recovering the surface geometry given a reference configuration in which the template shape is known, and a set of 3D-to-2D correspondences between this shape and the input image. On top of this, additional constraints enforcing isometry~\cite{salzmann2008closed}, conformal warps~\cite{bartoli2015shape} and photometric consistency~\cite{moreno2013stochastic,moreno2009capturing} are considered. While effective, SfT methods are very sensitive to the initial set of  matches, which may be difficult to establish in practice, especially under occlusions, low textured surfaces  and varying illumination.

Temporal information is another typically exploited prior. Non-rigid-shape-from-motion techniques generally extend Tomasi and Kanade's rigid factorization algorithm~\cite{tomasi1992shape} to recover deformable shape and camera motion from a sequence of 2D tracks, exploiting physical~\cite{AgudoCVPR2015} and low-rank constraints on the shape~\cite{AgudoJMIV2016,AgudoCVPR2017,LeeCVPR:2014,TorresaniPAMI2008}, trajectory~\cite{AkhterPAMI2011} or the forces inducing the deformation~\cite{AgudoICCV2015}. Again, these methods rely on the fact that 2D point tracks can be readily computed, limiting thus their general applicability to relatively well-textured surfaces.

The need of point correspondences is circumvented by template-free approaches that perform a per-point 3D reconstruction by minimizing an objective function on geometric and photometric cues~\cite{balan2007shining,gorce2008model,wang2016template,white2006combining}. The shading models considered by these approaches, however, use to be oversimplifications of the reality, either considering brightness constancy~\cite{wang2016template} or Lambertian surfaces lit by point light sources~\cite{white2006combining}.

More realistic deformation and appearance models can be learned from training data. The first attempt along this line corresponds to the active appearance models~\cite{cootes2001active}, which learned low-dimensional 2D models for face tracking. This was later extended to 3D by the active shape and morphable models~\cite{Blanz:etal:1999,matthews2004active}, and by methods integrating these models into the SfT formulation~\cite{moreno2011probabilistic}. Yet, all these approaches still rely on feature points detected over the whole surface or at its boundary~\cite{salzmann2008local}, which are difficult to obtain in practice. 

Following the success of recent deep convolutional networks in related topics such as 3D human pose recovery~\cite{martinez2017simple,moreno20163d,pavlakos2016coarse}, depth~\cite{eigen2015predicting,eigen2014depth,garg2016unsupervised,liu2016learning,roy2016monocular,xu2017multi} and surface normal reconstruction on rigid objects~\cite{bansal2017pixelnet,bansal2016marr,eigen2015predicting,wang2015designing}, we introduce a unified formulation for the problem of estimating non-rigid shape from single images, that simultaneously performs 2D detection and 3D lifting while enforcing geometry consistency. The framework we propose allows tackling a series of situations which, to the best of our knowledge, are not jointly addressed by existing approaches for reconstructing deformable surfaces: it does not require pre-computing point correspondences, it is effective on poorly textured surfaces, it is robust to partial occlusions and corrupted object boundaries, and works well under varying lighting conditions. Moreover, 3D shape inference is fast as often with deep networks.

Probably the most closely related work to ours is that of Tewari~\etal~\cite{tewari2017mofa}, which trains a deep auto-encoder model for monocular face reconstruction. However, this work relies on a low-rank shape model that limits their feasible solutions to shapes with relatively small   deformations. Furthermore, the range of textures for face reconstruction is limited while we consider general textures.

\section{Our Approach}

Our framework for estimating a non-rigid shape from a single image is shown in Fig.~\ref{fig:network}. We have devised an architecture with three  branches, each responsible of reasoning about a different geometric aspect of the problem. The first two branches are arranged in parallel and perform probabilistic 2D detection of the mesh in the image plane and depth estimation (red and green regions in the figure, respectively). These two branches are then merged (blue region in the figure) in order to lift the 2D detections to 3D space, such that the estimated surface correctly re-projects onto the input image and it is properly aligned with the ground truth shape. In the results section we will show that reasoning in such a structured way provides much better results than trying to directly regress the shape from the input image, despite using considerably deeper networks. 

\section{Geometry-Aware Network}
In this section we formulate the problem and describe the network architecture we propose, which is made of three main branches named 2D Detection Branch, the Depth Branch, and the Shape Branch. We also define the loss layer for learning the whole model.

\subsection{Problem Formulation}
We aim at designing a deep learning framework that directly estimates a non-rigid 3D shape from an input RGB image $\bI\in\mathbb{R}^{H_o\times W_o \times 3}$. The  shape is represented as a triangulated 3D mesh with $N_v$ vertices $\bX=(\bx_1,\ldots,\bx_{N_v})$, where $\bx_i=(x_i,y_i,z_i)$ are the coordinates of the $i$-th vertex, expressed in the camera coordinate system. In the following, we assume the structure of the mesh to be known, being a $N\times N$ rectangular grid, i.e., $N_v=N^2$. 

We also assume the calibration parameters of the camera to be known, namely the focal lengths, $f_u$ and $f_v$, and the principal point $(u_c,v_c)$.

\subsection{2D Detection Branch}
Given an input image $\bI$, the first step consists in extracting image features from a pre-trained network, in our case we concatenate two Resnet V2 blocks~\cite{he2016identity}. For each block, the stride of the last unit is set to one, in order to keep the same spatial resolution for the two units. Let us denote these features as $\Psi(\bI)\in\mathbb{R}^{H\times W\times C}$.

\begin{figure}[t!]
	\includegraphics[trim={0 0 0 00mm},clip,width=0.47\textwidth]{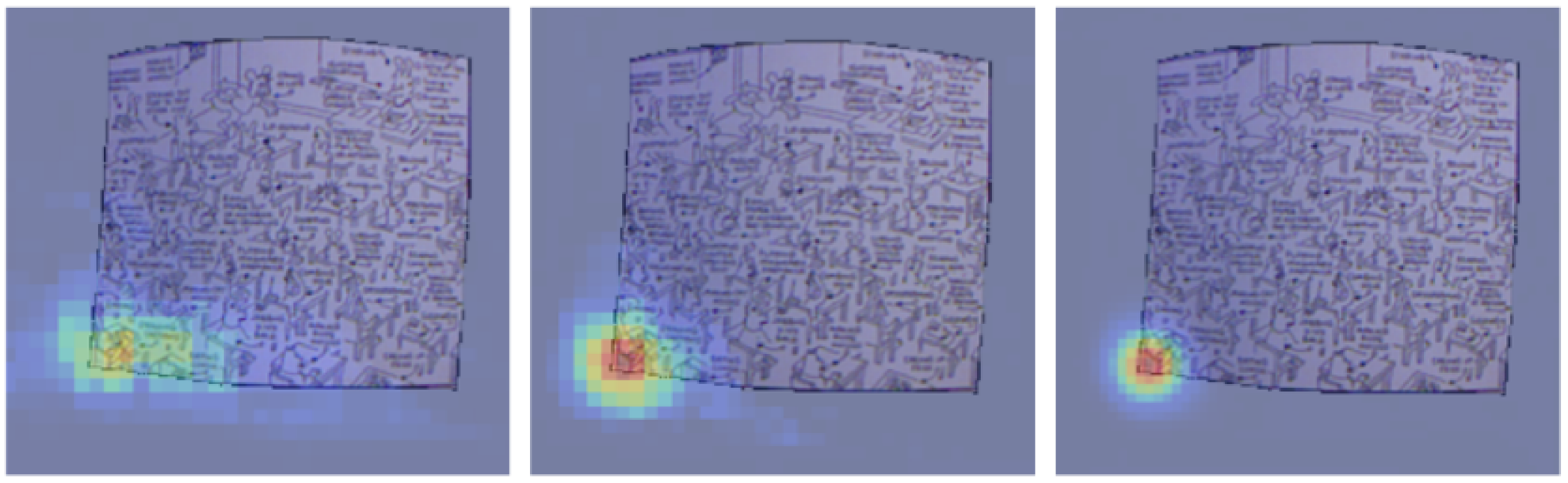}	
	\begin{tabular}{p{3cm} p{3cm} p{3cm}}
		\hspace{1cm}t=1& \hspace{0.2cm} t=2 & \hspace{-0.5cm} t=3
		\end{tabular}
	\caption{{\bf Refinement of the 2D vertices position. } Output (for one specific vertex) of the regressor $\Phi^t$  for three consecutive time steps. Note how the uncertainly in the vertex location is progressively reduced. }
	\label{fig:pose_g}	
\end{figure}

The image features are then fed into the 2D detection network, which is responsible for estimating the 2D locations of the mesh vertices  $\bU=(\bu_1,\ldots,\bu_{N_v})\in \mathcal{U}$, where $\bu_i=(u_i,v_i)$ and  $\mathcal{U}$ is the set of all $(u,v)$ pixel locations in the input image $\bI$. Drawing inspiration on the convolutional pose machines~\cite{wei2016cpm} for human pose estimation, the 2D location of each vertex $\bu_i$ is represented as a probability density map $\bB_i\in \mathbb{R}^{H\times W}$ computed over the entire image domain as:
\begin{equation}
\bB_i [u,v]= P(\bu_i = (u,v))\;\;, \forall \; (u,v)\in\mathcal{U} .
\end{equation}
 As in~\cite{wei2016cpm} these belief maps are estimated in an iterative manner. In particular, let $\bB^t=(\bB_1^t,\ldots,\bB_{N_v}^t)\in\mathbb{R}^{H\times W \times N_v}$  be the concatenation of all belief maps at iteration $t$. This tensor is estimated by a regressor function $\Phi^t$, which takes as input the image features and the concatenated belief maps at the previous stage $t-1$:
 \begin{equation}
 \Phi^t(\Psi(\bI),\bB^{t-1}) \rightarrow \bB^t \;.
 \end{equation}
In the first step, the regressor is only fed with the image features, that is $\Phi^{1}\equiv \Phi^{1}(\Psi(\bI))$. We denote by $T_{\textrm{max}}$ the maximum number of iterations. As it is shown in Fig.~\ref{fig:pose_g}, after each iteration, the location of the vertices is progressively refined.

In order to implement the regressor $\Phi^t(\cdot)$ we use again ResNet V2 blocks followed by two convolutional layers. The output of each  $\Phi^t$ is normalized with respect to $H$ and $W$  to guarantee that $\sum_{u=1}^{H}\sum_{v=1}^{W}\bB_i^t[u,v]=1$,  $\forall i \in \{1,\ldots,N_v\}$, and $\forall t \in \{1,\ldots,T_{\textrm{max}}\}$.

Finally, it is worth noting that the 2D Detection Branch we have just described is fully differentiable. The output $\bu_i=(u_i,v_i)$ for the $i-$th vertex can be estimated as the following weighted sum over the last belief map $\bB^{T_{\textrm{max}}}$:
\begin{equation}\label{eq:detection2d}
u_i\hspace{-1mm}=\hspace{-1mm}\frac{\sum\limits_{(u,v)\in\mathcal{U}}u\cdot \bB^{T_{\textrm{max}}}_i[u,v]}{\sum \bB^{T_{\textrm{max}}}_i}, \hspace{5mm} %
v_i\hspace{-1mm}=\hspace{-1mm}\frac{\sum\limits_{(u,v)\in\mathcal{U}}v\cdot \bB^{T_{\textrm{max}}}_i[u,v]}{\sum \bB^{T_{\textrm{max}}}_i}
\nonumber
\end{equation}
where $\sum \bB^{T_{\textrm{max}}}_i$  sums over all elements of $\bB^{T_{\textrm{max}}}_i$. These  2D estimates will be forwarded to the `Shape Branch' described in Section~\ref{sec:lifting}, while the belief maps in $\bB^{T_{\textrm{max}}}$ will be used to infer the depth value for each of the vertices in the `Depth Branch' described in Section~\ref{sec:depthbranch}.

\subsection{Depth Branch}  \label{sec:depthbranch}

The belief maps $\bB_i^{T_{\textrm{max}}}$ of the 2D vertex locations in the above section are forwarded to the `Depth Branch', to estimate the depth coordinate $z_i$ for every vertex. Note that previous works in related problems like 3D human pose estimation~\cite{martinez2017simple,moreno20163d} have not taken advantage of the uncertainty typically associated to the feature detectors.

To do so, the proposed layer produces new feature maps $\bV(\bB^{T_{\textrm{max}}} ,\Psi(\bI))\in \mathbb{R}^{N\times N \times C}$, that condition the input feature maps $\Psi(\bI)\in \mathbb{R}^{H \times W \times C} $ with the probability maps  $\bB^{T_{\textrm{max}}}\in \mathbb{R}^{H\times W \times N_v}$, that is:
\begin{equation}
\bV[j(i),k(i),c] =\sum_{(u,v)\in\mathcal{U}} \bB_i^{T_{\textrm{max}}}[u,v]\cdot \Psi(\bI)[u,v,c]
\end{equation}
$\forall i \in\{1,\ldots,N_v\}, c\in\{1,\ldots,C\}$,  where $(j(i),k(i))$ converts the $i$-th input of an $N_v$-dimensional vector into a two dimensional input of an  $N\times N$  matrix (recall that $N_v=N^2$).

These image features conditioned on the vertices 2D locations are then used as input of a regressor $\Omega(\cdot)$ to estimate the vertices' depth:
\begin{equation}
\Omega(\bV(\bB^{T_{\textrm{max}}},\Psi(\bI))) \rightarrow (z_1,\ldots,z_{N_v}).
\label{eq:depth}
\end{equation}
Again, the regressor $\Omega(\cdot)$ consists in two ResNet V2 blocks followed by two convolutional layers and the full branch (conditioned features + regressor) is fully differentiable.

\subsection{Shape Branch} \label{sec:lifting}
The 2D locations and depth estimates are merged in order to estimate the shape while enforcing the projection constraints and rigid alignment consistency.

Given the estimates $(u_i, v_i,z_i)$ in Eqs.~\eqref{eq:detection2d} and~\eqref{eq:depth} of the two first branches, the 3D position $\bx_i= (x_i,y_i,z_i)$ of each vertex is recovered with a differentiable layer that models the pinhole reprojection model:
\begin{equation}
x_i=z_i\cdot\frac{u_i-u_c}{f_u}, \hspace{7mm} y_i=z_i\cdot\frac{v_i-v_c}{f_v}, \hspace{7mm} z_i = z_i \> .
\end{equation}
This gives us an estimate of the deformable shape $\bX$,   and we could train the network by considering the L2 loss $|| \bX - \bX^*||_2^2$ where $\bX^*$ is the ground truth 3D shape. However, we propose introducing an additional layer, which computes the  Procrustes alignment error between $\bX$ and $\bX^*$ in a fully differentiable manner, and build our loss function based on this error. Although this layer is removed at test time, we observed that it favors the convergence during training, helps adapting to different datasets, and most importantly, it improves the capacity of the rest of the network to capture the non-rigid component of the shape. 

The Procrustes layer (`Procr' box in Fig.~\ref{fig:network}) is implemented by first normalizing   $\bX$ and $\bX^*$ with respect to translation and scale. Let us denote by $\hat{\bX}=(\hat{\bx}_1,\ldots,\hat{\bx}_{N_v})$ and $\hat{\bX}^*=(\hat{\bx}_1^*,\ldots,\hat{\bx}_{N_v}^*)$ these normalized versions.

Following~\cite{coutsias2004using}, we can then compute the alignment error between $\hat{\bX}$ and $\hat{\bX}^*$, without having to explicitly estimate their relative rotation and translation as follows:
\begin{equation}
\textrm{Err\_Align}(\hat{\bX},\hat{\bX}^*) = \sqrt{\frac{\sum_{i=1}^{N_v} |\hat{\bx}_i|^2 + |\hat{\bx}_i^*|^2 - 2\lambda_{\textrm{max}}   }{N_v}}
\label{eq:alignerror}
\end{equation}
where $\lambda_{\textrm{max}}$ is the maximum eigenvalue of a $4\times 4$ matrix built in terms of the elements of $\hat{\bX}$ and $\hat{\bX}^*$.  Since there exist differentiable approximations of the eigendecomposition (for example, the function \texttt{tf.self\_adjoint\_eigvals} in Tensorflow), the full `Shape branch' is again differentiable.

\subsection{Learning the Model}
The cost function that we aim to minimize is a combination of the 3D alignment error in Eq.~\eqref{eq:alignerror} and the 2D detection error produced at the output of each regressor $\Phi^t$, for $t=\{1,\ldots,T_{\textrm{max}}\}$:
\begin{equation}
\mathcal{L}=\textrm{Err\_Align}(\hat{\bX},\hat{\bX}^*)+ \gamma \sum_{t=1}^{T_{\textrm{max}}}\|\bB^t-\bB^* \|_2^2 \> ,
\label{eq:loss}
\end{equation}
where $\bB^*$ is a heat-map generated by placing Gaussian peaks at the ground truth 2D locations $(u_i^*,v_i^*)$ of the mesh vertices. $\gamma$ denotes a weight used to give similar orders of magnitude to each of the terms of the loss function. 

\vspace{1mm}
\noindent{\bf Training Details.} The model is trained with the synthetically generated dataset described in the next section, made of $H_o \times W_o = 224 \times 224$ images. 
The image features $\Psi(\bI)$ are obtained from a Resnet V2 network pre-trained on ImageNet, resulting in feature maps of size $H \times W \times C = 56 \times 56 \times 768$. In all our experiments we consider meshes of spatial resolution $N\times N = 9 \times 9$, thus, $N_v= 81$. The resulting belief maps $B^t$ will be therefore of size $56 \times 56 \times 81$. In the `2D Detection Branch', we fixed the maximum number of iterations to $T_{\textrm{max}}=3$, as further stages did barely change the resulting belief maps distributions.

   	    \begin{figure*}[t!]
  	    	\centering
  	    	\begin{tabular}{l}
  	    		\begin{tabular}{p{2.5cm}  p{2.5cm} p{2.5cm} p{2.6cm} p{2.6cm} p{2.6cm}    }
  	    			{\hspace{4mm}\small \bf Known Texture} &  {\hspace{5mm}\small \bf New Texture}  &{\hspace{4mm}\small \bf Non-Textured }& {\small \bf Known Texture}& {\hspace{1mm}\small \bf New Texture }  & {\small \bf Non-Texture}  \\
                      	    			{\hspace{4mm}\small \bf } &  {\hspace{5mm}\small \bf }  &{\hspace{4mm}\small \bf }& {\small \bf with Occlusion}& {\hspace{1mm}\small \bf with Occlusion }  & {\small \bf with Occlusion}  \\
  	    		\end{tabular}\\
  	    		{\includegraphics[trim={0 0 0 14mm},clip,width=\textwidth]{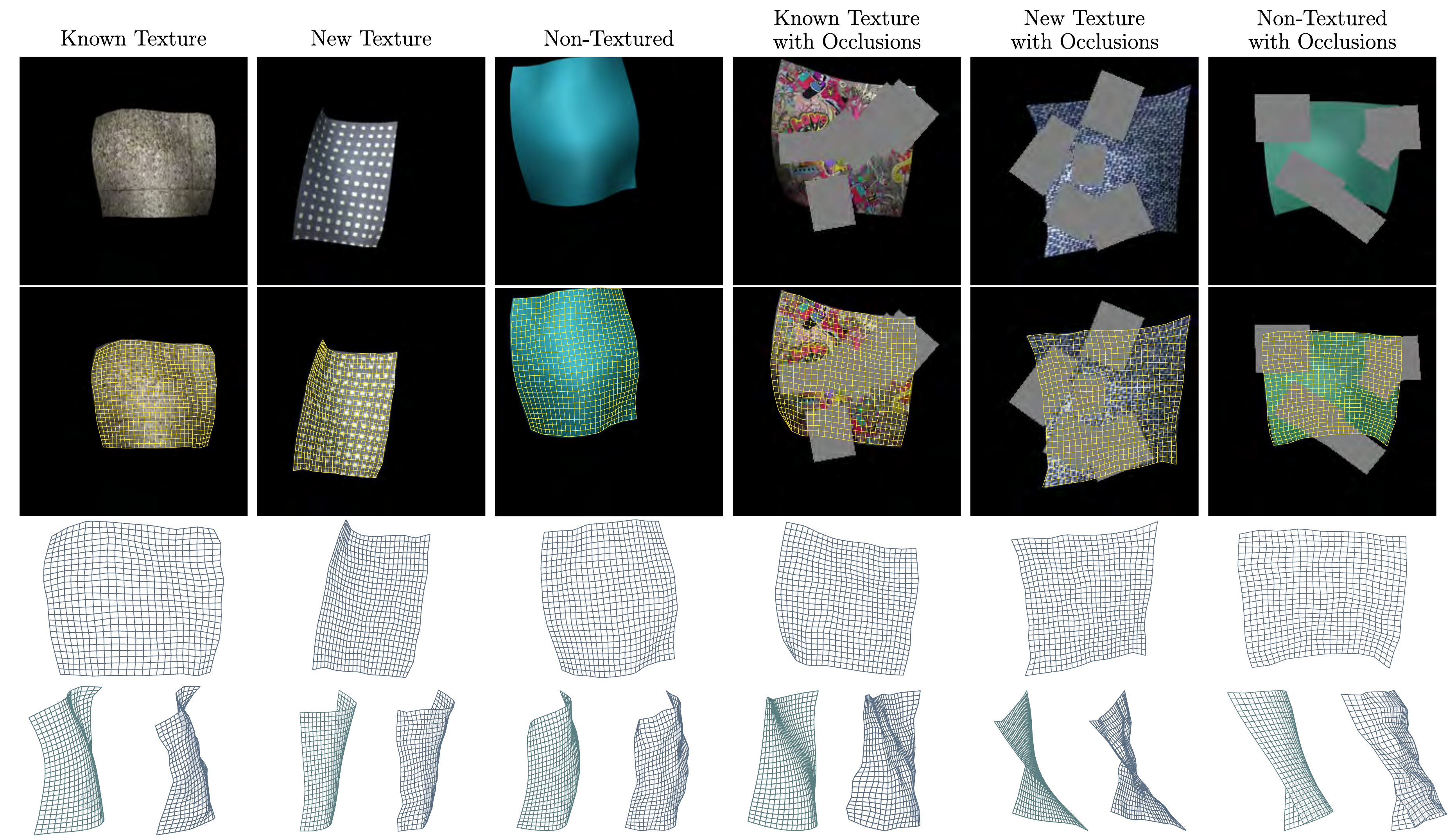}}\\
  	    		\begin{tabular}{p{2.5cm}  p{2.5cm} p{2.5cm} p{2.5cm} p{2.5cm} p{2.5cm}    }
  	    			{\hspace{10mm}2.42}  &  {\hspace{10mm}3.15}  & {\hspace{10mm}2.90}  & {\hspace{10mm}3.28}  & {\hspace{10mm}2.88} & {\hspace{10mm}2.83}\\
  	    		\end{tabular}
  	    	\end{tabular}
  	    	\vspace{-2mm}
  	    	\caption{{\bf Results on synthetic data.} Reconstructions samples in each of the six cases we consider (surfaces with known, new or no-texture, and with and without occlusions). {\bf First Row:} Input image. {\bf Second Row:} 3D estimated mesh projected onto the input image. {\bf Third Row:} 3D estimated mesh seen from the camera view. {\bf Last Row:} Side view of the ground truth mesh and our estimation (green and blue meshes, respectively). The reconstruction error is indicated at the bottom, to give significance to the errors in Table~\ref{tab:Tablesynthetic}.}
            \vspace{-4mm}
  	    	\label{fig:syntheticresults}
  	    \end{figure*}

The training procedures is split in two stages: initially, only the regressors $\Phi^t$ are trained. Then, regressors $\Phi^t$ and $\Omega$ are jointly trained. In both cases, the parameters of the feature extractor $\Psi(\bI)$ are kept fixed. In Eq.~\eqref{eq:loss} we set $\gamma = 5\cdot 10^{-3}$. We use Adam solver~\cite{kingma2014adam} with a batch size of 3 images and weight decay of $4\cdot 10^{-5}$. Every 2 epochs we exponentially decay the learning rate, which is initially set to $2\cdot 10^{-4}$.

\section{Dataset}
\label{sec:Dataset}
It is well known that deep networks require large amounts of training data. However, the only existing dataset we are aware of that contains non-rigid surfaces annotated with ground-truth 3D shape is~\cite{varol2012constrained}, which includes 505 images of a bending paper and a deforming t-shirt. This is far below what is needed, specially if we expect our network to generalize to non-observed textures. For this purpose, we have created a large synthetic dataset with 128,000 samples rendered with Autodesk\texttrademark - Maya. Each sample consists of a $224 \times 224$  image and a $9\times 9$ deformed shape. A few examples of the dataset are shown in Fig.~\ref{fig:syntheticresults}-Top.

We generated our dataset by varying textures, deformations and lighting conditions. Concretely, we have chosen 200 different textures from~\cite{synthesizability} which is formed by repetitive patterns, rich, poor and plain textures. The deformations were generated for 40 different meshes (same topology but varying aspect ratios and sizes). The mesh dynamics were rendered by simulating a hanging piece of material held with up to 4 pins and moving with the wind. Four different materials, defined with four different stiffness matrices, were considered. The scene was lit by one point light source of high intensity with a random position, plus a component of ambient illumination. In all cases, we assumed a Lambertian reflectance.

The rendered dataset was augmented with all three possible flips of each image. Additionally, for each image, three new ones were generated by applying a random rigid transformation on the corresponding deformable surface. At training time, the dataset was further augmented with random color changes at pixel level (hue, saturation, contrast and brightness). The dataset will be made publicly available.

\begin{table}[t!]
	\setlength{\tabcolsep}{3pt} 
	\setlength\arrayrulewidth{0.9pt}
	\centering
	\resizebox{8.3cm}{!} {
		\begin{tabular}{|l |c|c|c|c|}
			\hline
			\multicolumn{1}{|l|}{\cellcolor[gray]{0.95} \bf Method} & \multicolumn{1}{c|}{\cellcolor[gray]{0.95} \bf Known Text}   & \multicolumn{1}{c|}{\cellcolor[gray]{0.95} \bf New Text} & \multicolumn{1}{c|}{\cellcolor[gray]{0.95} \bf No-Text} & 
			\multicolumn{1}{c|}{\cellcolor[gray]{0.95} \bf Time (ms)} \\			
			\hline\hline
			Ba15Iso &  8.54 / - &  8.72 / - &   - / - & 495	 \\
			Ba15Iso-It &  5.65 / - &  6.78 / - &   - / - & 15,507	 \\
			Ba15Conf &  30.50 / - &  31.91 / - &  - / - & 11,232	 \\
			Ch14IsoLsq &  6.74 / - &  6.95 / - &  - / - & 2618	 \\
			Ch14IsoLsq-It &  4.85 / - &  5.3 / - &  - / - & 14,813	 \\
			Resnet-50 V2 &  {\bf 0.92} / {\bf 3.83} &  11.23 / 18.50 &  8.39 / 9.43 &  {\bf 152}	 \\
			DeformNet &  2.64 / 4.57 &  {\bf 3.28} / {\bf 4.09} &  {\bf 2.86} / {\bf 4.62}  &  219	 \\
			
			\hline
		\end{tabular}}
		\vspace{1.0mm}
		\caption{\textbf{Evaluation on synthetic data.} Euclidean average distance between 3D ground-truth and estimated 3D reconstruction. Each pair `err1 / err2' indicates the error without and with occlusions, respectively. Execution time in the last column is computed as the average time (in ms) to reconstruct a sample. Symbol `-' indicates that the method was not evaluated on this scenario, as they correspond to situations (no texture or large occlusions) that can not be addressed by template-based analytical solutions.} \label{tbl:Synth} \vspace{-2mm}
		\label{tab:Tablesynthetic}
	\end{table}

\section{Experimental Validation}  
We now present results on synthetic and real data. 
We compare our approach, which we dub \emph{DeformNet}, with the following state-of-the-art template-based solutions: Ba15Iso, the isometry-based solution proposed in~\cite{bartoli2015shape}; Ba15Conf, a conformal-based  approach, also from~\cite{bartoli2015shape}; Ch14IsoLsq, the least-squares isometric reconstruction of~\cite{chhatkuli2014stable}. We denote by Ba15so-It and Ch14IsoLsq-It the same previous methods after executing 25 iterations of the non-linear refinement proposed in~\cite{brunet2010monocular}. This refinement step could not be applied to Ba15Conf due to computational time constraints. \cite{chhatkuli2014stable} showed that Ch14IsoLsq-It systematically outperformed the same baselines we consider here and also the methods introduced in \cite{brunet2010monocular,perriollat2011monocular,salzmann2009reconstructing}. We therefore consider Ch14IsoLsq-It to be the best current analytic approach to assess the potential of our solution. Additionally, we also compare against a deep network baseline, consisting of a ResNet-50 V2 architecture~\cite{he2016identity} directly inferring 3D mesh coordinates.

In the following, we will report the reconstruction error, computed as the L2 distance between the estimated and the ground truth shapes (dimensionless for the synthetic results and in mm for the real ones). As common practice, the estimated meshes are aligned to the ground truth before evaluation using a Procrustes transformation. Additionally, in order to make a fair comparison, all methods requiring the pixels coordinates of the mesh, are fed with the estimates $\bU=(\bu_1,\ldots,\bu_{N_v})$ obtained with our network, augmented to a few hundreds of template-to-image correspondences by  interpolation. We would like to point that our network produces an error of approximately 2 pixels in these 2D detections, and computing them using feature descriptors such as SIFT~\cite{LoweIJCV2004}, generally led to worse results as these type of descriptors are prone to fail for non-textured surfaces with repetitive pattens and self-occlusions.

\subsection{Evaluation on Synthetic Data}

 We evaluated all methods on a test set of our dataset consisting of 1208 independent samples generated with random values of   shape and camera pose. These test samples are split into three subsets: 553 unknown shapes with a texture seen at training time (`Known Texture'), 553 unknown shapes with a texture not seen at training time (`New Texture'), and 102 unknown shapes without texture or very poorly textured (`Non-Textured'). Additionally we have simulated occlusions by covering the input images with a number of gray rectangular patches randomly distributed. Examples of the type of input images for each test case are shown in Fig.~\ref{fig:syntheticresults}-Top.

Template-based analytical methods  (Ba15Iso, Ba15Conf, Ch14IsoLsq and their iterative versions) were only evaluated on the textured and non-occluded cases, as they are methods that by construction can not realistically address the lack of texture or strong occlusions.   Alternatively, to make the learning approaches (Resnet-50 V2 and  DeformNet) robust to occlusions, the two networks were retrained with the `gray-patched' images. No retraining was done to handle the lack of texture. 
 
Table~\ref{tab:Tablesynthetic} summarizes the results of the synthetic evaluation. When dealing with textured and non-occluded images, Ch14IsoLsq-It is, as expected, the most accurate solution among the analytical methods. Regarding the learning approaches, Resnet-50 V2 turns to work very well under known textures. However, its performance suffers a big drop when dealing with textures not seen during training and with poorly textured surfaces.   DeformNet performs consistently well in all situations, outperforming in all cases the analytical solutions. Particularly interesting is the case when dealing with new textures that are occluded, in which we obtain an accuracy very similar to the best analytical methods (we obtain 3.62mm versus 3.57mm for competing methods) when dense non-occluded correspondences are provided.

Figure~\ref{fig:syntheticresults} shows examples of the reconstructed meshes obtained by our approach. Note that when there are no occlusions, the recovered shape highly resembles the ground truth, even for non-textured surfaces and not previously seen textures. When the input image is corrupted by occlusions, our solutions turn to be noisier, but even in this case, they are very close to the ground truth.
 
\vspace{1mm}
\noindent{\bf Computation Times.}
Another advantage of  learning based approaches is that once they are learned, they are much faster than the analytical solutions. The last column of Table~\ref{tab:Tablesynthetic} shows that computing the shape can be done in a fraction of a second for either Resnet-50 V2 and our approach, between one and two orders of magnitude faster than analytical methods. 

\begin{figure*}
	\resizebox{17.3cm}{!} {
		\begin{tabular}{@{}ccc@{}}
			\includegraphics[trim={35mm 90mm 40mm 100mm},clip,width=0.35\linewidth]{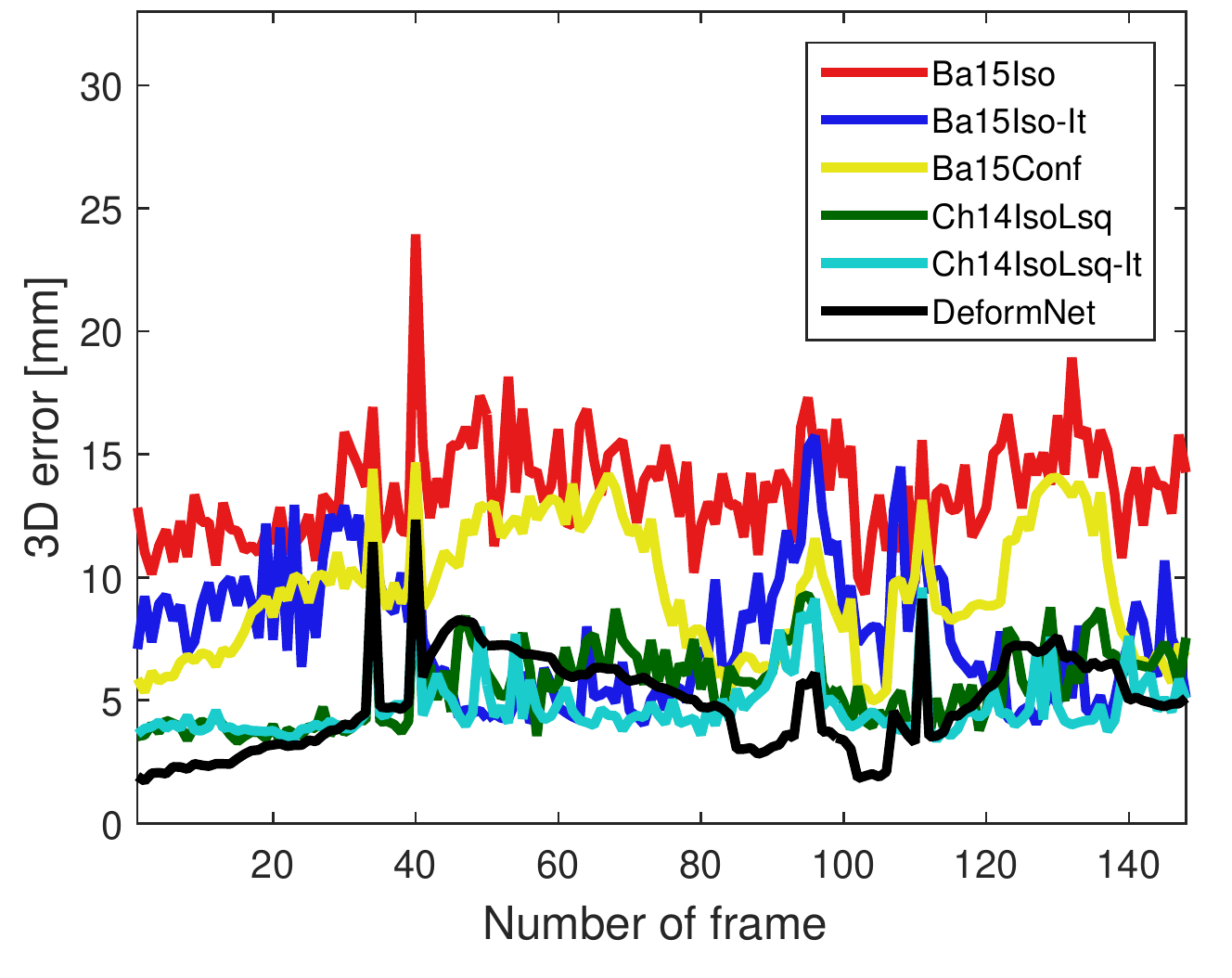}&
			\hspace{-0.51cm}
			\includegraphics[trim={35mm 90mm 40mm 100mm},clip,width=0.35\linewidth]{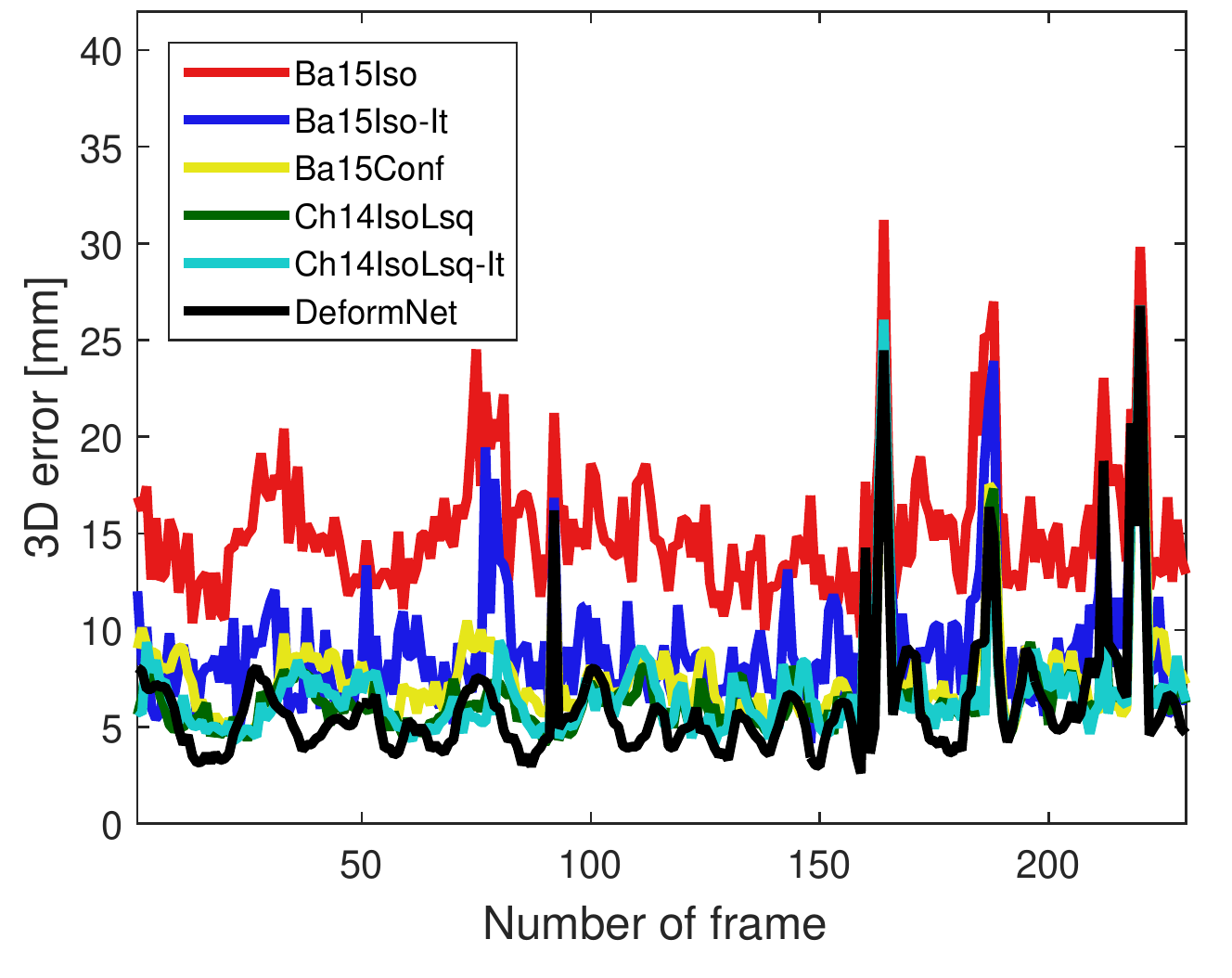}&
			\includegraphics[trim={5mm 49mm 5mm 50mm},clip,width=0.3, width=0.3\linewidth]{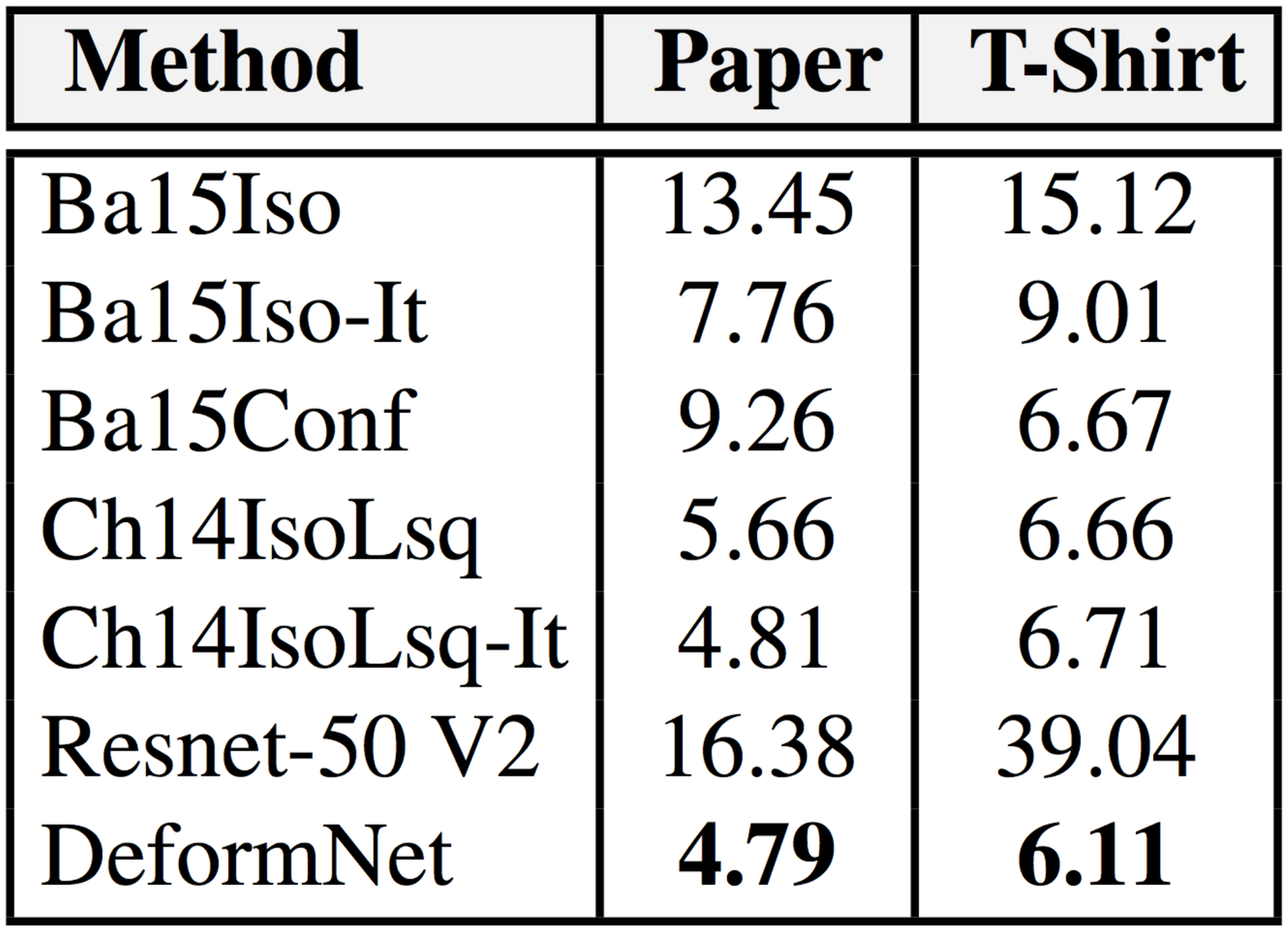}			
		\end{tabular}}
		\vspace{-0.2cm}
		\caption{{\bf Evaluation on the  CVLab sequences~\cite{varol2012constrained}.} The two graphs plot the 3D reconstruction error per frame (in mm) for all methods in   the two real sequences (Left: Paper  bending sequence, Right: T-shirt sequence). The results of Resnet-50 V2 are not plotted as it   was not able to generalize to these sequences. {\bf Right.} Mean reconstruction errors of all methods.}
		\label{fig:seq_real_data}
		\vspace{-0.2cm}
	\end{figure*}
    
	  \begin{figure*}[h!]
  	\includegraphics[trim={0 0 0 00mm},clip,width=\textwidth]{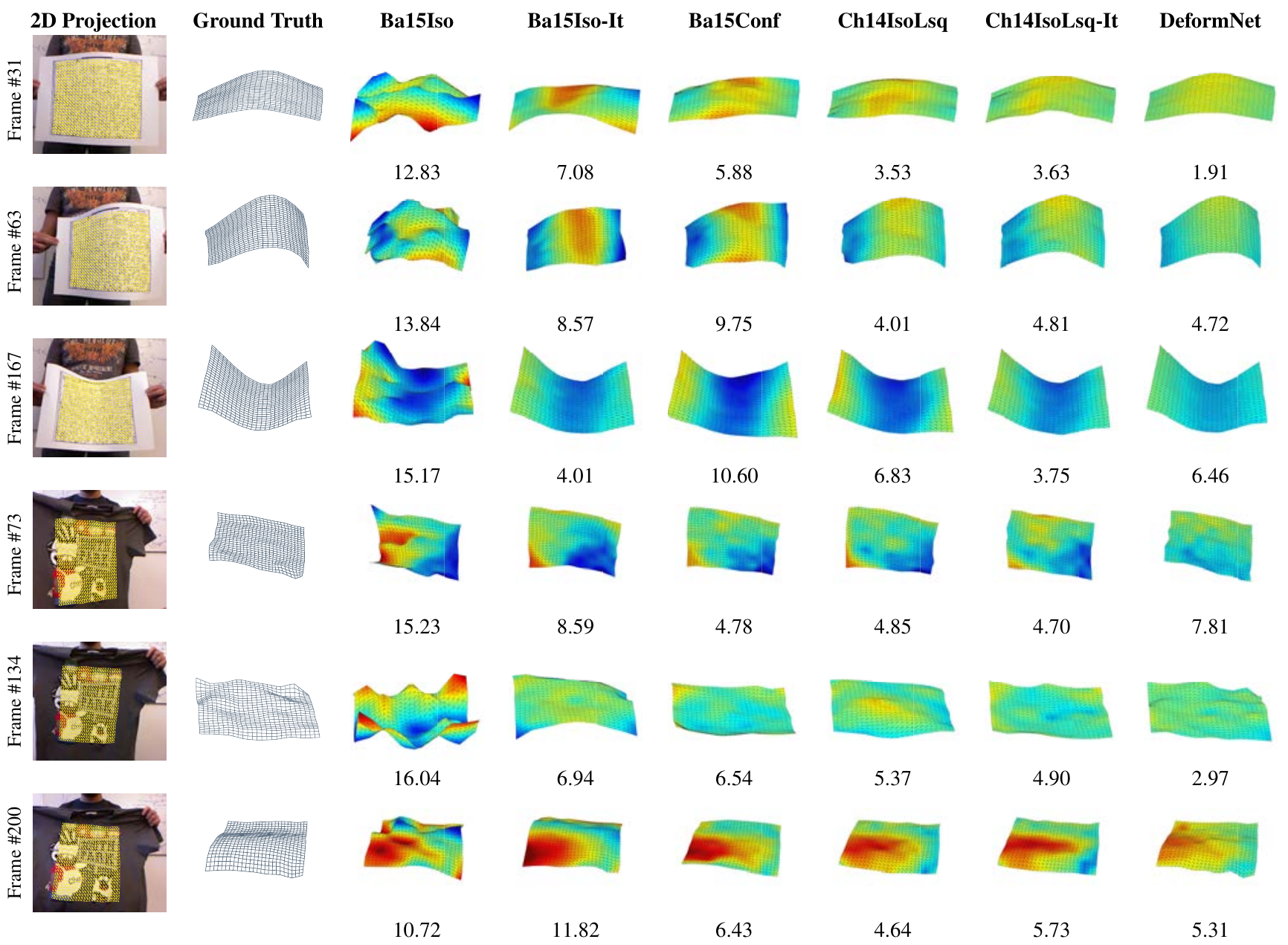}		
  \caption{{\bf Reconstructed meshes on the `paper bending' and `t-shirt'  CVLab sequences. } Results on Resnet-50 are not included as it did not generalize to real sequences.  Each shape is color coded according to its reconstruction error. Larger errors appear in red, and small errors in dark blue. Below each reconstructed shape we indicate the mean reconstruction error (in mm). }
  \vspace{-4mm}
  \label{fig:realsequences}	
  \end{figure*}

\subsection{Evaluation on Real Data} 

We also   evaluate all methods on two real datasets provided by CVLab~\cite{varol2012constrained}, which consist in video sequences of a bending paper and a deforming t-shirt, with  193 and 312 frames, respectively. As common practice, the background of the sequences was subtracted. Additionally, both for Resnet-50 V2 and DeformNet, we performed a finetuning of the networks with a very small portion of the dataset (15\% first frames). This finetuning was necessary to capture the bounds of the real deformations and adapt to the true illumination conditions that were not rendered by the synthetic dataset. In all methods we evaluated with the rest of the 85\% of the frames. Again, for the fairness of comparison,  the analytical solutions were fed by the 2D inputs of the mesh obtained by DeformNet, augmented to 500 correspondences using interpolation. The mean 2D location error (in pixels) obtained using DeformNet was 1.24 (paper bending sequence) and 2.28 (t-shirt sequence).

In Fig.~\ref{fig:seq_real_data} we plot the  3D reconstruction error per frame for all methods.  The table on the right of the figure  summarizes the results. Again, our DeformNet is the most accurate approach. In the bending paper sequence the analytic solution of Ch14IsoLsq-It is very close to ours, although DeformNet improves this method by a larger margin in the t-shirt sequence.  In any event, recall that DeformNet performs inference per image in a fraction of a second while Ch14IsoLsq-It requires about 15 seconds. For these sequences, Resnet-50 V2, the other deep learning baseline we considered, performs very poorly demonstrating that the specific architecture we use in DeformNet allows for a much better generalization.

Finally,   Fig.~\ref{fig:realsequences} shows a few reconstructed shapes obtained for each of the methods. Below each sample, we indicate the reconstruction errors. Note that samples with errors of about 4mm (in the paper bending sequence) or 6mm (in the t-shirt sequence) are already very good solutions. This is the magnitude of the error obtained by DeformNet.

\subsection{Discussion}
One of the most  significant aspects of our network is its ability to generalize to unknown textures (see results in Table~\ref{tab:Tablesynthetic}). We conjecture that this is the result of two factors: 1) training with a large variety of textures, and 2) separating the network into two input branches, one for performing 2D detection and the other to modulate input image features using the belief maps of the 2D detections. That is, our two branches allow us to correctly combine appearance and geometry. Note that the Resnet-50 V2 baseline we evaluated was also trained with a variety of textures, but it was not capable to generalize to new textures.

It is well known that on developable surfaces one may reconstruct shape from only the image boundaries~\cite{gumerov2004structure}. One might therefore think that the robustness of DeformNet to new textures might  be because our architecture learns to infer shape from the boundaries. In order to evaluate this, we performed the following experiment.

\begin{figure}[t!]
  	\includegraphics[width=\linewidth]{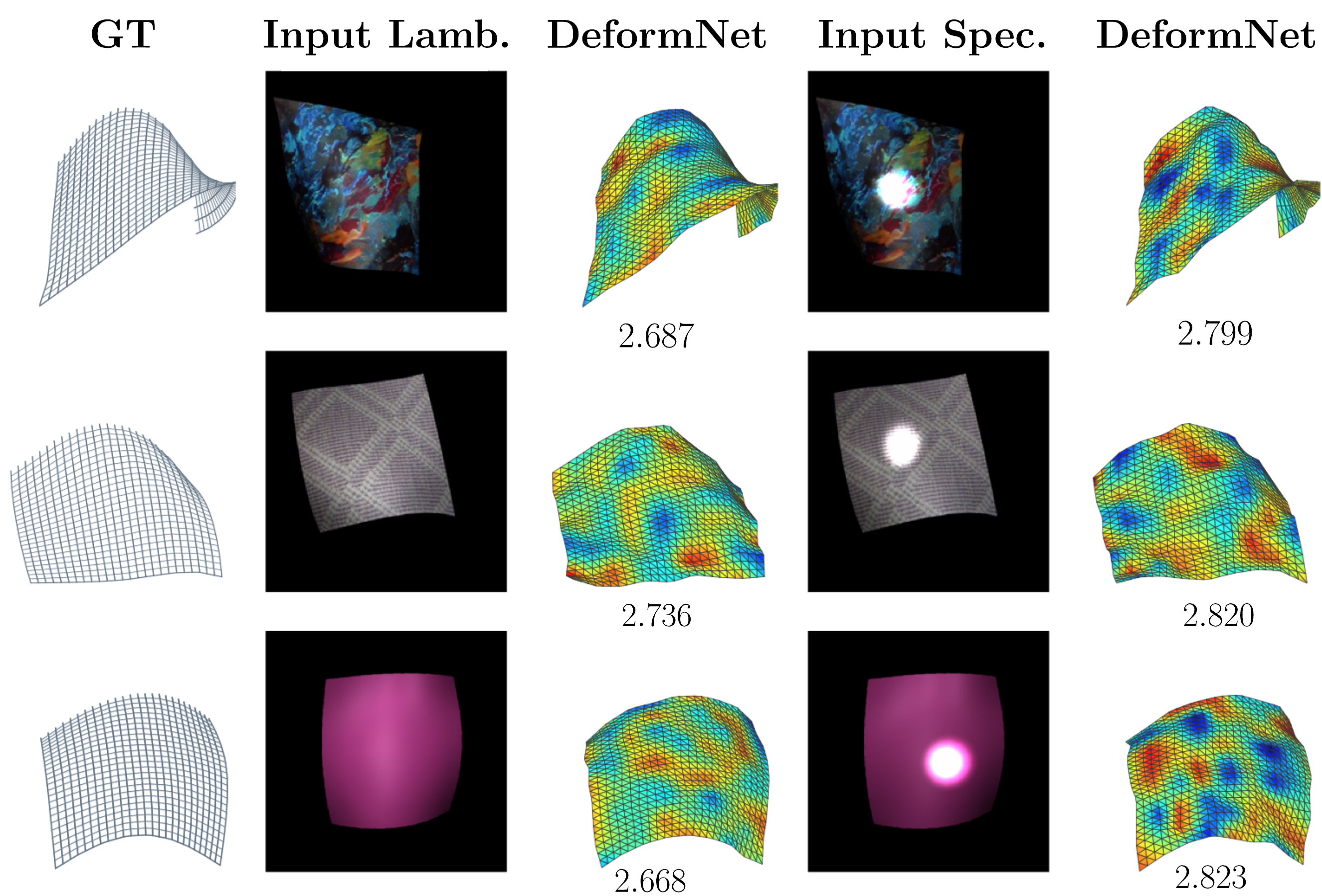}		
  \caption{{\bf Reconstruction under artificial specularities.} As in  Fig.~\ref{fig:realsequences}, each shape is color coded according to its reconstruction error.}
  \label{fig:lambertian}	
  \vspace{-2mm}
\end{figure}

\vspace{1mm}
\noindent{\bf Blurred contours.} In order to lower the dependency of DeformNet on the contours, we retrained it on a training set in which the surface boundaries of the input images were artificially corrupted by both adding random noise to the 2D coordinates of the boundary vertices and then blurring the contours. This strategy was also used   in~\cite{liberknecht2009dataset} to evaluate planar homographies.  We then tested our architecture on the full dataset and obtained an error of 3.77mm, which is just slightly above the results reported in Table~\ref{tab:Tablesynthetic}. Therefore, we can conclude that our network does not highly depend on the boundaries and exploits the whole image.

\vspace{1mm}
\noindent{\bf Relaxing Lambertian reflectance assumptions.} To further test our model limits Fig.~\ref{fig:lambertian} presents an evaluation of the model under synthetic specularities. The network also shows robustness to this scenario, and the overall reconstruction error (2.82) remains very similar to the case with Lambertian assumptions. 

\section{Conclusion}
We have proposed the first deep network that estimates the 3D shape of a non-rigid surface from a single image. For this purpose we have designed an architecture that can be trained in an end-to-end manner, but that internally splits the problem in three stages: 2D detection, depth estimation and shape inference. The three stages are intimately connected and are executed by ensuring the satisfaction of geometric constraints such as correct 3D-to-2D reprojection and 3D-to-3D alignment between the estimated and the ground truth shapes. In order to train this network, we have rendered a large synthetic dataset of shapes under different levels of deformation, varying textures, material properties and illumination conditions. We have shown this network to outperform existing analytical solutions while being much more efficient, being able to tackle situations with large amounts of occlusion and very poorly textured surfaces. As part of future work, we aim at extending this solution to more complex deformations and further exploring the connections of our solution with analytic photometric methods. 
 
\vspace{1mm} 
\noindent{\bf Acknowledgments:} This work is supported in part by a Google Faculty Research Award, by the Spanish Ministry of Science and Innovation under projects HuMoUR TIN2017-90086-R, ColRobTransp DPI2016-78957 and  Mar\'ia de Maeztu Seal of Excellence MDM-2016-0656; and by the EU project AEROARMS ICT-2014-1-644271. We also thank Nvidia for hardware donation.

\balance
{\small
\bibliographystyle{ieee}
\bibliography{egbib}
}

\end{document}